\documentclass[runningheads]{llncs}

\usepackage{cite}
\usepackage{adjustbox}
\usepackage{graphicx}
\usepackage{adjustbox}
\usepackage{multirow}
\usepackage{float}
\usepackage{hyperref}
\usepackage{comment}
\usepackage{amsfonts}
\usepackage{mathtools}
\usepackage[table,xcdraw]{xcolor}

\begin{document}
\title{Denoising Architecture for Unsupervised Anomaly Detection in Time-Series \thanks{The final authenticated publication is available online at \url{https://doi.org/10.1007/978-3-031-15743-1_17}}}
\titlerunning{Denoising Architecture for Unsupervised Anomaly Detection in Time-Series}
\author{Wadie Skaf \orcidID{0000-0002-4298-6694} \and
Tomáš Horváth\orcidID{0000-0002-9438-840X}}

\authorrunning{W. Skaf and T. Horváth}

\institute{
Telekom Innovation Laboratories, Data Science and Engineering Department (DSED), Faculty of Informatics, Eötvös Loránd University, Pázmány Péter stny. 1/A, 1117, Budapest, Hungary. \\
\email{\{skaf, tomas.horvath\}@inf.elte.hu}}
\maketitle             
\begin{abstract}

Anomalies in time-series provide insights of critical scenarios across a range of industries, from banking and aerospace to information technology, security, and medicine. However, identifying anomalies in time-series data is particularly challenging due to the imprecise definition of anomalies, the frequent absence of labels, and the enormously complex temporal correlations present in such data. The LSTM Autoencoder is an Encoder-Decoder scheme for Anomaly Detection based on Long Short Term Memory Networks that learns to reconstruct time-series behavior and then uses reconstruction error to identify abnormalities. We introduce the Denoising Architecture as a complement to this LSTM Encoder-Decoder model and investigate its effect on real-world as well as artificially generated datasets. We demonstrate that the proposed architecture increases both the accuracy and the training speed, thereby, making the LSTM Autoencoder more efficient for unsupervised anomaly detection tasks.

\end{abstract}
\section{Introduction}
An outlier or anomaly is a data point that differs dramatically from the rest of the data. Hawkins \cite{Douglas1980} defined an anomaly as an observation that deviates significantly from the rest of the observations, raising suspicions that it was generated by an unusual mechanism. Anomaly detection is used in a variety of industries, including network intrusion detection, credit card fraud detection, sensor network malfunction detection, and medical diagnosis \cite{Chandola2009}.

In time-series data, an outlier or anomaly is a data point that deviates significantly from the overall trend, seasonal or cyclical pattern of the data. By significance, the majority of data scientists mean statistical significance, which indicates that the data point's statistical properties are out of phase with the rest of the series. Anomalies are classified into two broad categories: a point anomaly is a single data point that has reached an abnormal value, whereas a collective anomaly is a continuous sequence of data points that are considered anomalous collectively, even if the individual data points are not. Anomaly detection methods can be classified into the following two broad methodologies:
\begin{enumerate}
    \item {Semi-supervised anomaly detection} models are designed to be trained exclusively on data that does not contain anomalies and then tested on samples containing anomalies and non-anomalies to determine their accuracy.
    \item {Unsupervised anomaly detection} models are designed to be trained on data containing a mixture of anomalous and non-anomalous data samples without specifying which are which, and then tested on samples containing both anomalies and non-anomalies to determine their accuracy.
\end{enumerate}

An Encoder-Decoder scheme for anomaly detection (EncDec-AD) based on Long Short Term Memory Networks (LSTM) was proposed in \cite{Malhotra2016}, in which the encoder learns a vector representation of the input time-series, the decoder uses this representation to reconstruct the time-series, and the reconstruction error at any subsequent time instance is used to compute the likelihood of an anomaly at that point. However, since EncDec-AD trains on only regular sequences, it is a semi-supervised learning-based anomaly detection system. In this paper, we extend this model with a denoising architecture and put it to the test for unsupervised anomaly detection.

The paper is structured as follows. We begin by formalizing and discussing the topic of unsupervised time-series anomaly detection, delving into the details of the anomaly detection process using LSTM Autoencoders. After that, we describe the the proposed denoising architecture and set up the experiments, before reporting and summarizing our major findings.

\section{Related Work}
Numerous anomaly detectors based on classic statistical models have been presented throughout the years (e.g., \cite{Konorn2008, Lu2009, Pincombe2005, Yaacob2010ARIMABN}, mainly time series models) for computing anomaly scores. Due to the fact that these algorithms often make simplistic assumptions about the application domain, expert assessment is required to choose an appropriate detector for every particular domain and then fine-tune the detector's parameters using the training data. According to \cite{Liu2015OpprenticeTP}, simple ensembles of these detectors, such as majority vote \cite{Fontugne2010} and normalization \cite{Shanbhag2009}, are also ineffective. As a result, these detectors are seldom used in practice.

To address the difficulties associated with algorithm/parameter tweaking for classic statistical anomaly detectors, supervised ensemble techniques such as EGADS \cite{Laptev2015} and Opprentice \cite{Liu2015OpprenticeTP} have been developed. They train anomaly classifiers utilizing user feedback as labels and traditional detector output as features. Both EGADS and Opprentice demonstrated promising results, but they depend substantially on high-quality labels, which is often not practical in large-scale applications. Additionally, using numerous conventional detectors to extract features during detection adds significant computing overhead, which is a practical problem.

Recently, there has been an increase in the use of unsupervised machine learning algorithms for anomaly detection, including one-class SVM \cite{Amer2013, ERFANI2016121}, clustering-based approaches such as K-Means \cite{Munz07trafficanomal} and Gaussian Mixture Model (GMM) \cite{Laxhammar2009AnomalyDI}, Kernel Density Estimation (KDE) \cite{Cao2016}, Auto-Encoder (AE) \cite{An2015}, Variational Auto-Encoder (VAE) \cite{An2015}. The aim is to place an emphasis on normal patterns rather than on anomalies. Generally speaking, these algorithms generate the anomaly score by first identifying "normal" regions in the original or some latent feature space, and then determining "how distant" an observation is from the normal regions.

Additionally, significant progress has been made lately in training generative models using deep learning approaches for the purpose of performing anomaly detection, such as Generative Adversarial Network (GANs) \cite{Li2019, Zhou2019, Yoon2019, Geiger2020}.

Despite the enormous potential of the aforementioned models and algorithms, SVM, KDE, AE, and VAE are not designed to handle time-series data, and using more complicated models such as GANs results in lengthy training times and high resource requirements. Given the aforementioned limitations, and given that deep learning architectures have exceptional learning abilities and are particularly adept at tolerating non-linearity in complicated temporal correlations \cite{Kwon2019}, we enhance the LSTM Encoder-Decoder architecture \cite{Malhotra2016} that has been used in supervised anomaly detection by introducing the Denoising Architecture, which enables it to perform unsupervised time-series anomaly detection in addition to significantly reducing the training time.

\section{Unsupervised Time-Series Anomaly Detection}

Given a time-series $X = (\mathbf{x}_{1}, \mathbf{x}_{2}, \dots, \mathbf{x}_{T})$, unsupervised time-series anomaly detection is the process of identifying segments $A^j=(\mathbf{x}_{j}, \mathbf{x}_{j+1}, \dots, \mathbf{x}_{j+n_j})$ of anomalous points, where $\mathbf{x}_{i} \in \mathbb{R}^N$, $j \geq 1$, $n_j \geq 0$ such that $j+n_j \leq T$, and $N=1$ in the case of a univariate time-series while $N>1$ in the case of a multivariate time-series. 
Each $A^j$ is a single data point (when $n_j = 0$) or a continuous series of data points (when $n_j > 0$) in time that exhibit(s) anomalous or unexpected behavior value(s) inside the segment that do not appear to conform to the signal's predicted temporal behavior.

This process is different from and is more complicated than time-series classification \cite{Fawaz2018} and supervised time-series anomaly detection \cite{Qiu2019} in a few aspects:
\begin{enumerate}
    \item {Absence of prior knowledge of anomalies or prospective anomalies}: unlike the supervised methods, which use previously recognized anomalies to train and optimize the model, the unsupervised methods use all the data to train the model to understand the time-series patterns, then ask it to find anomalies. The detector will then be checked to ensure that it recognized anything useful to end users. Additionally, $n_j$ (the length of $A^j$) is variable and unknown in advance, complicating this method even further.
    \item {Unsupervised methods do not rely on baselines}: for a large number of real-world systems, simulation engines can generate a signal that approximates normal contexts, providing a baseline against which models can be trained, with any deviations considered anomalies. Unsupervised time-series anomaly detection algorithms do not depend on such baselines but rather learn time-series patterns from real data that may contain anomalies or abnormal patterns.
    \item {Not all identified anomalies are cause for concern}: detected anomalies may not necessarily indicate issues, but may be the result of an external phenomena such as a rapid change in ambient circumstances, additional information such as a test run, or other factors not evaluated by the system, such as regime settings changes. In this context, it is up to the end user or domain expert to determine whether the anomalies detected by the model are worrisome.
\end{enumerate}

\section{Anomaly Detection using LSTM Autoencoder}
Autoencoder is a generative unsupervised deep learning model that reconstructs high-dimensional input data by utilizing a neural network with a narrow bottleneck layer that contains the latent representation of the input data in between the Encoder and Decoder. The autoencoder attempts to minimize reconstruction error as part of its training procedure. As a result, the magnitude of the reconstruction loss can be used to detect anomalies.

During the training process, the data is transferred to the Encoder, which generates a fixed-length vector representation of the input time-series. This representation is then used by the LSTM decoder to reconstruct the time-series using the current hidden state and the estimated value at the previous time step. Given a time-series of length $L$ as an input, $X = (\mathbf{x}_{1}, \mathbf{x}_{2}, \dots, \mathbf{x}_{T})$, where $\mathbf{x}_{i} \in \mathbb{R}^{N}$, $E_k^{(t)}$ is the hidden state of the $k^{th}$ layer of the encoder at time $t \in \{1, \ldots, L\}$, where $k \in \{1 \ldots H\}$; $H$ denotes the number of hidden layers in each of the encoder and the decoder, and $E_k^{(t)} \in \mathbb{R}^u$; $u$ is the number of LSTM units in hidden layer $k$ of the encoder. The final state of the final hidden layer $E_H^{(L)}$ of the encoder outputs the data's latent representation, which is used as an initial state of the decoder. The decoder then reconstructs the original input by using the input $x^{(i)}$ to obtain the hidden state $D_1^{(i-1)}$ (The hidden state of the first layer of the decoder at time ($i-1$)) then proceed with all the hidden states of the first layer then outputs $z^{(i)}_1$ to the next hidden layer, and so on until the final layer, when the decoder utilizes $z_{H-1}^{(i)}$ to to derive the hidden state $D_H^{(i-1)}$ then estimates $\mathbf{x}^{\prime}_{i-1}$ corresponding to $\mathbf{x}_{i-1}$. The autoencoder is trained with the purpose of minimizing the following: 
\begin{equation}
    \sum_{X} \sum_{i=1}^{L} \left \| \mathbf{x}_{i} - \mathbf{x}^{\prime}_{i} \right \|^2
\end{equation}

After training the autoencoder, time-series signals $I = \{X_1, X_2, ...., X_N\}$ of length L are passed to it in order to reconstruct them, then the reconstruction error of each point is calculated using equation \ref{eq:reconstruction_errr}
\begin{equation}
\label{eq:reconstruction_errr}
    s_i = \| \mathbf{x}_{i} - \mathbf{x}^{\prime}_{i}\|
\end{equation}
 where $\mathbf{x}_{i} \in X_i; X_i \in I$ can be used as an anomaly score, and by specifying a reconstruction error threshold, anomalous values can be flagged if the reconstruction error (equation \ref{eq:reconstruction_errr}) exceeds the value specified threshold .

\section{The Denoising Architecture}
Dropout \cite{srivastava2014dropout} is a strategy for decreasing overfitting in neural networks. Backpropagation learning by itself accumulates brittle co-adaptations that work for the training data but do not generalize to unobserved data. By making the existence of any specific hidden unit unstable, random dropout disrupts these co-adaptations. This approach was discovered to significantly increase the performance of neural networks across a broad range of application fields. The term "dropout" refers to units that are dropped from a neural network (both hidden and visible). By dropping a unit from the network, we imply disconnecting it from all incoming and outgoing connections temporarily.\\
In our proposed architecture, we add a dropout layer after each LSTM layer in the LSTM Autoencoder. As a result, during the training phase, the output of each LSTM layer would be randomly set to 0 with a probability $p$, which is done by generating a random number $r$, and if this number is less than or equal to $p$, the output would be set to zero, otherwise it would pass with no changes, as shown in equation \ref{eq:denoising-eq}.

\begin{equation}
\label{eq:denoising-eq}
    Output(x) =
    \begin{cases*}
    0 & if $r \leq p$ \\
    LSTM(x) & Otherwise
    \end{cases*}
\end{equation}

The proposed denoising architecture randomly exposes the model to extreme cases (zeroes) during training, allowing it to more accurately generalize the normal samples without being significantly affected by anomalous samples, so that its weights do not change significantly in the presence of an anomalous sample.  As a result, after training, the model will be capable of efficiently reconstructing normal samples while struggling to reconstruct anomalous samples, resulting in a higher reconstruction errors (equation \ref{eq:reconstruction_errr}) for the anomalous samples, allowing for a more precise definition (threshold) of anomalies. The selection of probability $p$ is based on a number of factors, which are discussed in section \ref{sec:results}.

\begin{figure}
    \centering
    \includegraphics[width=0.8\linewidth]{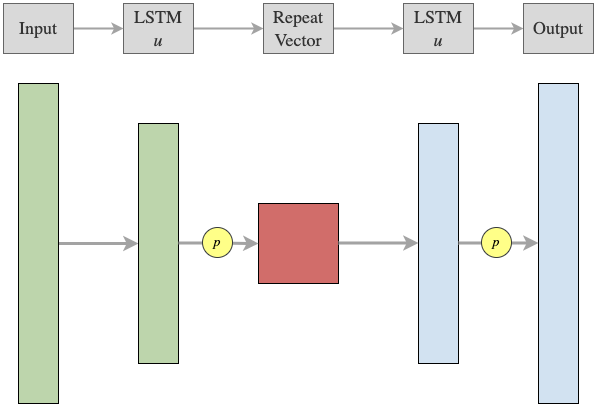}
    \caption{An illustration of an LSTM autoencoder with two LSTM layers, each has $u$ units and followed by a dropout layer with a probability of $p$}
    \label{fig:denoising_architecture_example}
\end{figure}

Figure \ref{fig:denoising_architecture_example} shows an example of an LSTM Autoencoder with a denoising architecture.

\section{Experimental Results}

\subsection{Datasets}
\label{section-datasets}
To examine the denoising architecture's effect on the LSTM Autoencoder, we use the Yahoo S5 dataset \footnote{Yahoo S5 Dataset can be requested here: \url{https://webscope.sandbox.yahoo.com/catalog.php?datatype=s&did=70}}, which is composed of four subsets: A1, A2, A3, and A4. The A1 dataset is based on real-world production traffic on Yahoo systems, whereas the remaining datasets are all made up of synthetic data. A2 lacks anomaly points and thus cannot be used to calculate the metrics described in section \ref{section-experimental-setup}; therefore, it is not used in the experiments. All values in the datasets are timestamped with a one-hour timestep.
\subsection{Experimental Setup}
\label{section-experimental-setup}

\begin{enumerate}
    \item Data Preparation:
    We performed data normalization on each dataset to ensure that the data was within the range $[-1,1]$. A sliding window with a window size of 24 (representing 24 hours) and a step size of 1 was used to generate the training samples, resulting in a sequence of 24 consecutive data points for each training sample. \\
    \item Architectures:
    We examine the effect of the denoising architecture on a variety of architectures, beginning with the simplest possible Autoencoder with two LSTM layers – one for the Encoder and one for the Decoder – and progressing deeper as shown in table \ref{tab:expirements_results}. \\
    \item Evaluation Metrics:
    \label{section-metrics}
     We use Recall, Precision, and F1-Score matrices to quantify the effect of the denoising architecture on model accuracy, as well as the number of epochs to quantify the training time speed. \\
    \item Comparison Baseline:
    For each pair of (Dataset, Architecture), we will first train the model without any dropout layers and use that as a baseline for comparison, and then gradually add dropout layers with increasing probability up to 0.5 and compare the results to the baseline.
\end{enumerate}

\subsection{Benchmarking Results}
\label{sec:results}

We outline the results of the experiments in table \ref{tab:expirements_results},where we list each architecture along with the obtained results. The notion for architectures used in the table is as follows: for the sake of simplicity, the architecture is represented by the layers of the encoder, so number 16 represents an autoencoder with two LSTM layers, each of which has 16 units, one for the encoder and the other for the decoder.

\begin{figure}[]
    \centering
     \includegraphics[width=0.9\linewidth]{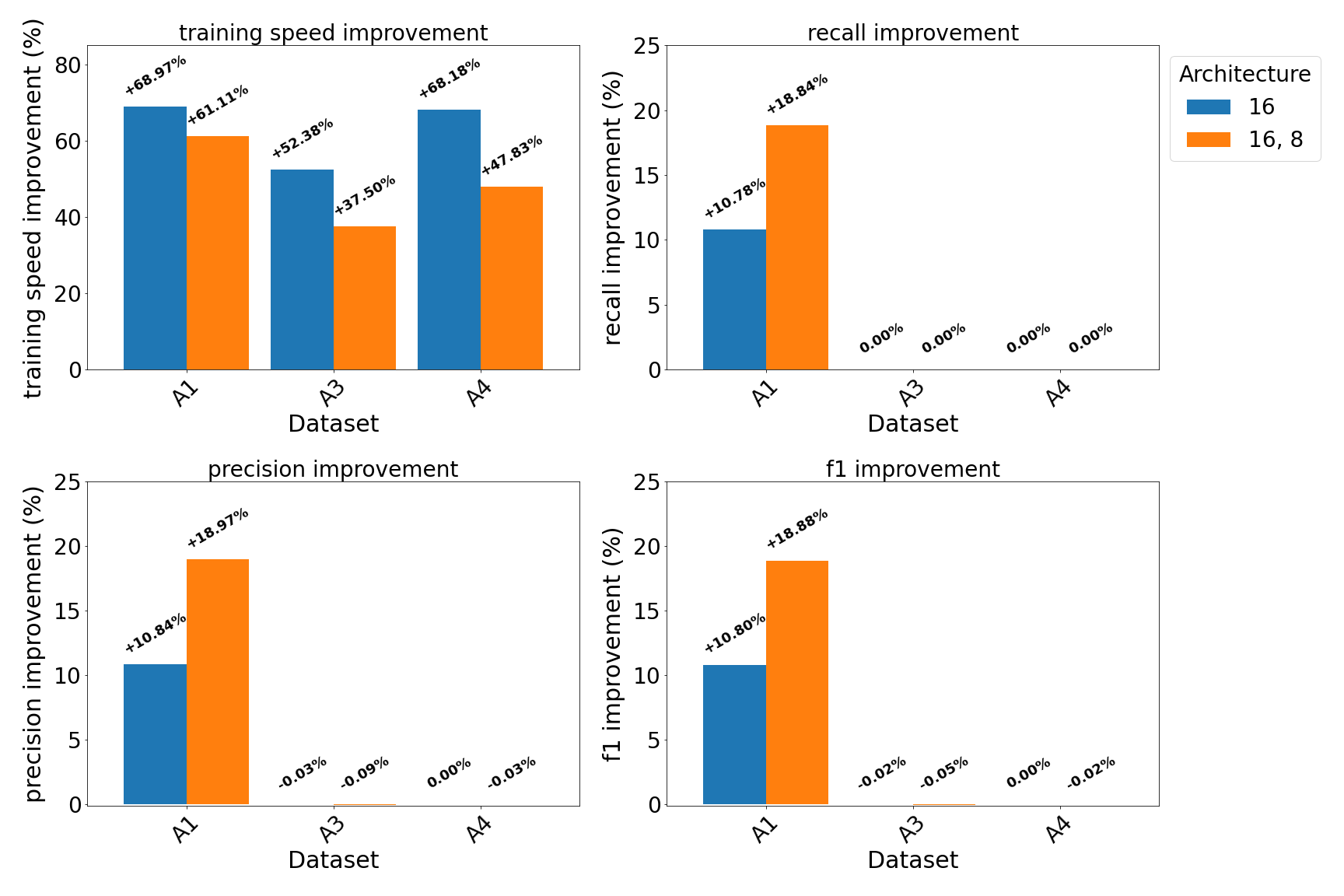}
    \caption{A comparison of the denoising architecture to the baseline, demonstrating the improvement in accuracy metrics and training speeds.}
    \label{fig:improvements}
\end{figure}

As shown in table \ref{tab:expirements_results}, \textbf{denoising improved the accuracy metrics on the real dataset (A1) and shortened the training time on both real and synthetic datasets (A1 \& A3 \& A4)}.
In comparison to the baseline (without dropout layers), improvements in recall, precision, f-1 score, and number of epochs are observed for the real dataset (A1), as well as the training speed for the all datasets (A1 \& A3 \& A4). All of these enhancements are outlined in table \ref{tab:expirements_results} and illustrated in figure \ref{fig:improvements}.In (A3 \& A4), a very slight decrease of $\epsilon < 0.001$ happened in precision and f-1 score , which can be ignored.

\begin{table}[h]
\renewcommand{\arraystretch}{1.2}
\caption{Experiments Results}\label{tab:expirements_results}
\centering
\begin{minipage}{0.5\linewidth}
\begin{adjustbox}{width=\linewidth}
\begin{tabular}{|c|cccccc|}
\hline
Dataset &
  \multicolumn{1}{c|}{Arch} &
  \multicolumn{1}{c|}{Dropout} &
  \multicolumn{1}{c|}{Epochs} &
  \multicolumn{1}{c|}{Recall} &
  \multicolumn{1}{c|}{Precision} &
  F1 \\ \hline
   &
   &
  0.0 &
  29 &
  0.198542 &
  0.442812 &
  0.274159 \\
 &
   &
  0.1 &
  14 &
  0.199482 &
  0.444911 &
  0.275459 \\
 &
   &
  0.2 &
  18 &
  0.205363 &
  0.457787 &
  0.283534 \\
 &
   &
  0.3 &
  18 &
  0.216184 &
  0.481656 &
  0.298425 \\
 &
   &
  \cellcolor[HTML]{C7FAC6}0.4 &
  \cellcolor[HTML]{C7FAC6}9 &
  \cellcolor[HTML]{C7FAC6}0.219948 &
  \cellcolor[HTML]{C7FAC6}0.490814 &
  \cellcolor[HTML]{C7FAC6}0.303769 \\
 &
  \multirow{-6}{*}{16} &
  0.5 &
  9 &
  0.217596 &
  0.485310 &
  0.300471 \\ \cline{2-7} 
  
   &
   &
  0.0 &
  18 &
  0.194778 &
  0.434190 &
  0.268918 \\
 &
   &
  0.1 &
  27 &
  0.209598 &
  0.466981 &
  0.289333 \\
 &
   &
  0.2 &
  11 &
  0.214067 &
  0.477189 &
  0.295551 \\
 &
   &
  0.3 &
  7 &
  0.220654 &
  0.492130 &
  0.304694 \\
 &
   &
  \cellcolor[HTML]{C7FAC6}{\color[HTML]{333333} 0.4} &
  \cellcolor[HTML]{C7FAC6}{\color[HTML]{333333} 7} &
  \cellcolor[HTML]{C7FAC6}{\color[HTML]{333333} 0.231475} &
  \cellcolor[HTML]{C7FAC6}{\color[HTML]{333333} 0.516535} &
  \cellcolor[HTML]{C7FAC6}{\color[HTML]{333333} 0.319688} \\
\multirow{-12}{*}{A1} &
  \multirow{-6}{*}{16, 8} &
  0.5 &
  14 &
  0.218772 &
  0.487933 &
  0.302095 \\ \hline
  \end{tabular}
  \end{adjustbox}
  
\end{minipage}%
\begin{minipage}{0.5\linewidth}
\centering
\begin{adjustbox}{width=\linewidth}
\begin{tabular}{|c|cccccc|}
\hline
Dataset &
  \multicolumn{1}{c|}{Arch} &
  \multicolumn{1}{c|}{Dropout} &
  \multicolumn{1}{c|}{Epochs} &
  \multicolumn{1}{c|}{Recall} &
  \multicolumn{1}{c|}{Precision} &
  F1 \\ \hline
   &
   &
  0.0 &
  21 &
  1.000000 &
  0.597741 &
  0.748233 \\
 &
   &
  0.1 &
  9 &
  1.000000 &
  0.597222 &
  0.747826 \\
 &
   &
  \cellcolor[HTML]{FFFFC7}0.2 &
  \cellcolor[HTML]{FFFFC7}10 &
  \cellcolor[HTML]{FFFFC7}1.000000 &
  \cellcolor[HTML]{FFFFC7}0.597568 &
  \cellcolor[HTML]{FFFFC7}0.748097 \\
 &
   &
  0.3 &
  18 &
  1.000000 &
  0.597568 &
  0.748097 \\
 &
   &
  0.4 &
  10 &
  0.990795 &
  0.591896 &
  0.741076 \\
 &
  \multirow{-6}{*}{16} &
  0.5 &
  9 &
  0.946221 &
  0.565268 &
  0.707737 \\ \cline{2-7} 
 &
   &
  0.0 &
  24 &
  1.000000 &
  0.597741 &
  0.748233 \\
 &
   &
  \cellcolor[HTML]{FFFFC7}0.1 &
  \cellcolor[HTML]{FFFFC7}15 &
  \cellcolor[HTML]{FFFFC7}1.000000 &
  \cellcolor[HTML]{FFFFC7}0.597222 &
  \cellcolor[HTML]{FFFFC7}0.747826 \\
 &
   &
  0.2 &
  12 &
  0.998062 &
  0.596583 &
  0.746783 \\
 &
   &
  0.3 &
  9 &
  0.928779 &
  0.555169 &
  0.694943 \\
 &
   &
  0.4 &
  14 &
  0.816376 &
  0.487699 &
  0.610618 \\
\multirow{-12}{*}{A3} &
  \multirow{-6}{*}{16, 8} &
  0.5 &
  14 &
  0.837209 &
  0.500000 &
  0.626087 \\ \hline
  \end{tabular}
\end{adjustbox}
  
\end{minipage} 

\begin{minipage}{0.5\linewidth}
\centering
\begin{adjustbox}{width=\linewidth}
\begin{tabular}{|c|cccccc|}
\hline
Dataset &
  \multicolumn{1}{c|}{Arch} &
  \multicolumn{1}{c|}{Dropout} &
  \multicolumn{1}{c|}{Epochs} &
  \multicolumn{1}{c|}{Recall} &
  \multicolumn{1}{c|}{Precision} &
  F1 \\ \hline
   &
   &
     0.0 &
  22 &
  1.000000 &
  0.401042 &
  0.572491 \\
 &
   &
  0.1 &
  11 &
  1.000000 &
  0.401042 &
  0.572491 \\
 &
   &
  \cellcolor[HTML]{FFFFC7}0.2 &
  \cellcolor[HTML]{FFFFC7}7 &
  \cellcolor[HTML]{FFFFC7}1.000000 &
  \cellcolor[HTML]{FFFFC7}0.401042 &
  \cellcolor[HTML]{FFFFC7}0.572491 \\
 &
   &
  0.3 &
  7 &
  1.000000 &
  0.401158 &
  0.572609 \\
 &
   &
  0.4 &
  10 &
  1.000000 &
  0.401042 &
  0.572491 \\
 &
  \multirow{-6}{*}{16} &
  0.5 &
  11 &
  1.000000 &
  0.401158 &
  0.572609 \\ \cline{2-7} 
 &
   &
  0.0 &
  23 &
  1.000000 &
  0.401390 &
  0.572846 \\
 &
   &
  0.1 &
  7 &
  1.000000 &
  0.401042 &
  0.572491 \\
 &
   &
  \cellcolor[HTML]{FFFFC7}0.2 &
  \cellcolor[HTML]{FFFFC7}12 &
  \cellcolor[HTML]{FFFFC7}1.000000 &
  \cellcolor[HTML]{FFFFC7}0.401274 &
  \cellcolor[HTML]{FFFFC7}0.572727 \\
 &
   &
  0.3 &
  14 &
  0.997114 &
  0.400116 &
  0.571074 \\
 &
   &
  0.4 &
  17 &
  0.994228 &
  0.398727 &
  0.569186 \\
\multirow{-12}{*}{A4} &
  \multirow{-6}{*}{16, 8} &
  0.5 &
  17 &
  0.805195 &
  0.323104 &
  0.461157 \\ \hline
 \end{tabular}
\end{adjustbox}
\end{minipage}
\end{table}

The optimal probability of dropout $p$ varies by dataset, being 0.4 for dataset A1 for all architectures, 0.2 for dataset A4 for all architectures, and 0.2 and 0.1 for dataset A3 for architectures (16), (16, 8), respectively. And to examine this, we investigate the ratio of anomaly points in each dataset presented to the model per epoch during training in the table \ref{tab:anomaly_samples_ratio}. As can be observed, there is a negative correlation between the number of anomaly samples and the optimal p, which sounds plausible because when there are more anomaly samples, the model will perceive more anomalies and will be more robust to them without requiring a higher dropout rate $p$ \textemdash The more anomaly samples the model perceives, the less a single anomaly sample significantly alters the neural network's weights. Thus, in general, the amount of $p$ should be determined by the number or ratio of anomaly samples, which in the case of unsupervised, can be determined by knowing the expected number of anomalies or by experimentation.

\begin{table}[h]
\caption{The percentage of anomaly points, the total number of samples, and the optimal $p (s)$ for each dataset}
\label{tab:anomaly_samples_ratio}
\centering
\begin{tabular}{c|c|c|c|}
\cline{2-4}
\multicolumn{1}{l|}{}                           & A1        & A3        & A4        \\ \hline
\multicolumn{1}{|c|}{Total number of samples}   & 2,238,624 & 3,974,400 & 3,974,400 \\ \hline
\multicolumn{1}{|c|}{Number of anomaly samples} & 6286      & 22,203    & 19,855    \\ \hline
\multicolumn{1}{|c|}{Anomaly samples percentage}     & 0.280\%   & 0.559\%   & 0.499\%   \\ \hline
\multicolumn{1}{|c|}{Optimal $p(s)$}                 & 0.4       & 0.1, 0.2  & 0.2       \\ \hline
\end{tabular}
\end{table}

\section{Conclusion}
In this paper, we introduced the Denoising Architecture as an addition to the LSTM Autoencoder to extend its usage to unsupervised anomaly detection for point anomalies and evidenced that it resulted in noticeable improvements in accuracy metrics such as precision, recall, and f-1 score (up to 18\%), as well as a remarkable increase in training speed (up to 68\%), and we demonstrated that improvements in accuracy occur only when real-world datasets are used, whereas synthetic datasets only show improvements in training speed. Additionally, we addressed how to choose the appropriate dropout probability $p$, showing that the more anomalous samples present or expected in the data stream, the smaller $p$ should be.

%
%
%
\bibliographystyle{splncs04}
%
\bibliography{citations}
\end{document}